\documentclass{article}

    \PassOptionsToPackage{numbers, compress}{natbib}

\usepackage[preprint]{neurips_2025}

\usepackage[utf8]{inputenc} 
\usepackage[T1]{fontenc}    
\usepackage{hyperref}       
\usepackage{url}            
\usepackage{booktabs}       
\usepackage{amsfonts}       
\usepackage{nicefrac}       
\usepackage{microtype}      
\usepackage{xcolor}         
\usepackage{amsmath}
\usepackage{amssymb}
\usepackage{mathtools}
\usepackage{amsthm}
\usepackage{colortbl}
\usepackage[capitalize,noabbrev]{cleveref}
\usepackage{multirow}
\usepackage{makecell}
\usepackage{siunitx}
\usepackage{subcaption}
\usepackage{graphicx}
\usepackage{pgfplots}
\usepackage{tikz}
\definecolor{brandeisblue}{rgb}{0.0, 0.44, 1.0}
\definecolor{brandeisbluecompl}{HTML}{FF8F00}
\definecolor{brandeis2}{HTML}{0F00FF}
\definecolor{brandeis4}{HTML}{00F0FF}
\definecolor{brandeispurple}{HTML}{6D00FF}
\usepackage{adjustbox}
\usepackage{tabularx}

\crefname{section}{Sec.}{Sec.}
\crefname{thm}{Thm.}{Theorem}
\crefname{appendix}{App.}{Appendices}
\crefname{algorithm}{Alg.}{Algorithms}
\crefname{equation}{Eq.}{Eqs.}
\crefname{figure}{Fig.}{Figs.}
\creflabelformat{equation}{#2\textup{#1}#3}

\usepackage{float}
\floatstyle{plaintop}
\restylefloat{table}
\usepackage[tableposition=top]{caption}

\newcommand{\myvec}[1]{\mbox{$\mathbf{#1}$}}
\newcommand{\myvecsym}[1]{\mbox{$\boldsymbol{#1}$}}
\newcommand{\vh}{\mbox{$\myvec{h}$}}
\newcommand{\vm}{\mbox{$\myvec{m}$}}
\newcommand{\vv}{\mbox{$\myvec{v}$}}
\newcommand{\vx}{\mbox{$\myvec{x}$}}
\newcommand{\vtheta}{\mbox{$\myvecsym{\theta}$}}
\newcommand{\vparam}{\vtheta}
\newcommand{\vA}{\mbox{$\myvec{A}$}}
\newcommand{\vB}{\mbox{$\myvec{B}$}}
\newcommand{\vW}{\mbox{$\myvec{W}$}}
\newcommand{\gauss}{\mbox{${\cal N}$}}
\newcommand{\dkls}[3]{\mathbb{D}_{\text{KL}}^{#1}[#2 \, \|\, #3]}
\newcommand{\loss}{\ell}
\newcommand{\bbE}{\mathbb{E}}
\newcommand{\vSigma}{\mbox{$\myvecsym{\Sigma}$}}
\newcommand{\param}{\theta}

\title{Improving LoRA with Variational Learning}

\author{%
  Bai Cong$^{1,2}$ \quad Nico Daheim$^3$ \quad Yuesong Shen$^{4}$ \\
  \textbf{Rio Yokota}$^1$ \quad \textbf{Mohammad Emtiyaz Khan}$^2$ \quad \textbf{Thomas Möllenhoff}$^2$ \\
  \\
  $^1$Institute of Science Tokyo \quad $^2$RIKEN Center for AI Project \\
  $^3$Ubiquitous Knowledge Processing Lab (UKP Lab), Department of Computer Science, \\ Technical University of Darmstadt \\
 National Research Center for Applied Cybersecurity ATHENE, Germany \\
  \quad $^4$Huawei Hong Kong Research Center \\
}

\begin{document}

\maketitle

\begin{abstract}
Bayesian methods have recently been used to improve LoRA finetuning and,
although they improve calibration, their effect on other metrics (such as accuracy) is marginal and can sometimes even be detrimental.
Moreover, Bayesian methods also increase computational overheads and require additional tricks for them to work well.
Here, we fix these issues by using a recently proposed variational algorithm called IVON.
We show that IVON is easy to implement and has similar costs to AdamW, and yet it can also drastically improve many metrics by using a simple posterior pruning technique.
We present extensive results on billion-scale LLMs (Llama and Qwen series) going way beyond the scale of existing applications of IVON.
For example, we finetune a Llama-3.2-3B model on a set of commonsense reasoning tasks and improve accuracy over AdamW by 1.3\% and reduce ECE by 5.4\%, outperforming AdamW and other recent Bayesian methods like Laplace-LoRA and BLoB.
Overall, our results show that variational learning with IVON can effectively improve LoRA finetuning.
\end{abstract}

\section{Introduction}
Low-Rank Adaptation (LoRA)~\citep{hu2021lora} is a widely-used parameter-efficient finetuning technique for large-scale pretrained models which enables finetuning billion-scale 
Large Language Models (LLMs) on a single consumer-grade GPU.
This has made it the go-to method for finetuning LLMs in settings with limited computational resources.
As a consequence, many new improvements have been proposed in various directions, such as parameter efficiency~\citep{he2022sparseadapter,ding2023sparse,zhang2023lora,kopiczkovera}, 
performance under quantization~\citep{dettmers2024qlora,xu2024qalora,liloftq}, and rank adaptation~\citep{zhang2023adaptive,ding2023sparse,valipour2023dylora,lialin2023relora}.

Recent works have also explored an alternative based on Bayesian methods to improve calibration. While this works well to some extent, there is a lot of room for improvement.
For example, Laplace-LoRA~\citep{yang2023bayesian} estimates a posterior that can be used for prediction by using Laplace's method on the LoRA parameters, but this requires 
additional post-hoc changes, including calculating a Kronecker-factored Hessian and model linearization. 
Despite increasing the overhead, only marginal improvements in accuracy are obtained. 
Another method called BLoB~\citep{wang2024blob} instead estimates the covariance during training using Bayes by Backprop~\citep{blundell2015weight}.
While it can improve generalization and calibration 
when using just the mean, the performance degrades when using posterior sampling.
BLoB also requires additional implementation tricks. For instance, uncertainty is only considered in one of the two LoRA blocks and flipout is used to introduce randomness~\citep{wen2018flipout}. 
Furthermore, BLoB increases the computation cost compared to standard non-Bayesian LoRA finetuning.
Ideally we would like a simpler alternative that can bring more benefits with less overhead. 

In this paper, we show that a recently proposed natural-gradient variational learning algorithm called IVON~\citep{shen2024variational} is a better alternative to improve LoRA with Bayesian principles.
IVON can simply be used to replace conventional optimizers like AdamW~\citep{loshchilov2018decoupled} as it shares a nearly identical implementation which makes it fast and easy to use.
While AdamW only provides a point estimate, IVON also estimates a diagonal Gaussian posterior over parameters during training.
This posterior also allows us to add cheap post-training pruning which drastically improves generalization and calibration.
These two aspects together comprise our method, which we call IVON-LoRA.

IVON-LoRA is easy to implement and achieves significant improvements for finetuning LLMs across various tasks and datasets.
For example, on a set of commonsense reasoning tasks
IVON-LoRA improves accuracy by 1.3\% and reduces calibration error by 5.4\% when compared to AdamW for finetuning Llama-3.2-3B, while outperforming other 
Bayesian-LoRA methods in terms of accuracy with comparable calibration. 
Accuracy and calibration can also be traded-off when sampling from the learned posterior with varying temperature.
Finally, we use the learned posterior for test-time compute scaling and improve math word problem solving accuracy on GSM8k with Qwen-2.5-3B.
Overall, we provide an easy-to-use improvement of LoRA using variational learning.

\begin{figure}[t]
  \centering
  \includegraphics[width=\linewidth]{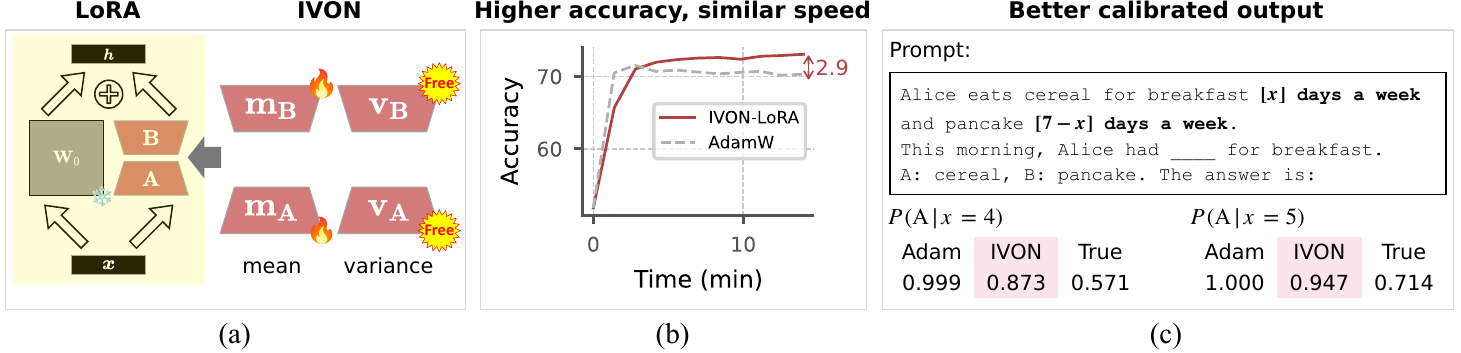}

  \caption{
    Overview and benefits of IVON-LoRA.
    (a) IVON-LoRA estimates a diagonal Gaussian posterior over LoRA parameters $(\mathbf{A}, \mathbf{B})$.
    IVON optimizes the means $(\vm_{\mathbf{A}}, \vm_{\mathbf{B}})$, while also yielding the variances ($\vv_{\mathbf{A}}, \vv_{\mathbf{B}}$) \textit{for free}.
    (b) IVON-LoRA converges as fast as AdamW but achieves significantly better accuracy (here we show an example on training the Llama-3.2-3B model on the WinoGrande-S dataset).
    (c) Given hand-crafted prompts that exhibit intrinsic uncertainty, LoRA adapters trained with IVON-LoRA assign probabilities that are better aligned with the ground truth calculated from the frequency mentioned in the prompt, indicating improved calibration.
  }
\end{figure}

\section{Efficient Finetuning with Low-Rank Adaptation}
The large sizes of current LLMs make it hard for many practitioners to finetune the full model due to resource constraints.
To overcome this, many parameter-efficient finetuning approaches have been proposed.
These methods usually either only train parts of the network, for example, the bias term~\citep{ben-zaken-etal-2022-bitfit}, or insert a small number of new parameters, for example, 
as prefixes to input token sequences~\citep{li-liang-2021-prefix} or at arbitrary positions in a model~\citep{pmlr-v97-houlsby19a, pfeiffer-etal-2020-mad, poth-etal-2023-adapters}.
However, these methods either exhibit poor performance or incur inference-time overhead~\citep{ruckle-etal-2021-adapterdrop}.

To tackle this, Low-Rank Adaptation (LoRA)~\cite{hu2021lora} inserts new parameters as a low-rank decomposition 
of the update applied to the original parameters during finetuning.
These new parameters can then be merged with the original parameters to not increase inference overhead.
Formally, given a weight matrix $\vW_0 \in \mathbb{R}^{d \times k}$ of a pretrained model, LoRA introduces a low-rank decomposition
\[
\Delta \vW_0 = \vB\vA,
\]
where $\vA \in \mathbb{R}^{r \times k}$, $\vB \in \mathbb{R}^{d \times r}$, and $r \ll \min(d, k)$.
The new weight matrix $\vW$ is then:
\[
\vW = \vW_0 + \Delta \vW_0 =\vW_0 + \vB\vA.
\]
The low-rank matrices $\vA$ and $\vB$ are initialized such that $\Delta \vW_0 = \vB\vA = 0$ to preserve the original $\vW_0$ at the beginning of training. 
During training, only $\vA$ and $\vB$ are optimized, while $\vW_0$ remains frozen. This drastically reduces the number of trainable parameters 
from $O(dk)$ to $O(2dr)$.
For linear layers, the forward pass with LoRA can be computed as:
\[
\boldsymbol{h} = \vW_0 \boldsymbol{x} + \Delta \vW_0 \boldsymbol{x} = \vW_0 \boldsymbol{x} + \vB\vA \boldsymbol{x},
\]
where $\boldsymbol{x} \in \mathbb{R}^{k}$ is the input and $\boldsymbol{h} \in \mathbb{R}^{d}$ the output.
For Transformers~\citep{VaSh17}, which are our main focus, LoRA is typically applied to the query and value matrices of the attention layers. 

Despite its computational efficiency, LoRA can lead to overconfident models~\citep{wang2023loraensembleslargelanguage} and lag behind full finetuning in terms of accuracy~\citep{biderman2024lora}.
To tackle this, Bayesian variants of LoRA have recently been proposed.
For example, Laplace-LoRA~\citep{yang2023bayesian} uses Laplace's method~\citep{mackay1992practical} to learn a posterior distribution over LoRA parameters by 
estimating the curvature around a point estimate $\vm$ trained with conventional learning algorithms like AdamW.
That is, $\vm$ contains all $\vA$ and $\vB$ matrices that are learned.
This results in a Gaussian posterior $q(\vparam) = \gauss(\vparam\mid\vm, \vSigma)$, where $\vSigma$ is the inverse of a Kronecker-factored (KFAC) approximation~\citep{pmlr-v37-martens15, ritter2018a} 
of the Fisher information matrix.
There are several problems with Laplace-LoRA though.
First, computing $\vSigma$ requires an additional pass through the training data which is not always available after training. An extra pass through the data also adds computational overhead.
Second, it requires a KFAC approximation of the Fisher information matrix which is another overhead.
Finally, prediction is done using a linearized model~\citep{pmlr-v130-immer21a},
but this requires the Jacobian $\nabla_{\text{\vparam}} f_{\text{\vparam}}(\vx)$ for the neural network outputs $f_{\text{\vparam}}(\vx)$.
This can be prohibitive for LLMs in a standard setting for next-token prediction due to requiring 
storage of $\mathcal{O}(d|\mathcal{V}|)$ for an output vocabulary $\mathcal{V}$ and number of parameters $d$.

A recently proposed method, BLoB~\citep{wang2024blob}, circumvents some of these issues by directly learning the mean $\vm$ and diagonal covariance $\vSigma$ during training with Bayes by Backprop~\citep{blundell2015weight}.
However, this requires multiple implementation changes to the standard LoRA.
For example, only $\vA$ is treated probabilistically and it is unclear how to prune $\vB$ based on the probabilistic information. Moreover, 
a new variant of flipout~\citep{wen2018flipout} is introduced. 
Altogether, this results in nontrivial implementation changes. 
In addition, both BLoB and Laplace-LoRA do not give large gains in accuracy over non-Bayesian LoRA finetuning and there is a trade-off between accuracy and calibration.
In \cref{sec:method,sec:ugp}, we propose a variational learning method that improves both accuracy and calibration of LoRA with minimal implementation changes.

\section{Variational Low-Rank Adaptation}
\label{sec:method}
Here we introduce our approach which we call IVON-LoRA.
The idea is straightforward: We replace the commonly used AdamW optimizer with IVON~\citep{shen2024variational} which optimizes a variational-Bayesian objective.
More formally, let us denote the AdamW objective by $\loss(\vparam)$ where $\vparam$ is the vector containing all entries of LoRA's low-rank parameters.
IVON-LoRA instead minimizes a Bayesian objective where an expectation of $\loss(\vparam)$ over a posterior distribution $q(\vparam)$ is used (shown on the right), 
\begin{equation}
   \min_{\text{\vparam}}  \,\, \loss(\vparam) \quad \text{vs.} \quad
   \min_{q(\text{\vparam})} \,\, \bbE_{q(\text{\vparam})} \left[ \loss(\vparam) \right] + \lambda^{-1} \, \dkls{}{q(\vparam)}{p(\vparam)}.
   \label{eq:AdamWvsIVON}
\end{equation}
IVON uses a diagonal Gaussian $q(\vparam) = \gauss(\vm, \text{diag}(\vv))$ with a zero mean isotropic Gaussian prior $p(\vparam)$ with a scalar variance.
The mean $\vm$ plays a similar role to $\vparam$ obtained by AdamW while the posterior variance $\vv$ captures additional information about 
the uncertainty over $\vm$ and enables sampling models from $q(\vparam)$.
The main advantage of our approach is that it only requires a few lines of training code to be changed,
thanks to the nearly identical implementation of IVON and AdamW.
The key point is that estimation of $\vv$ is done automatically through the scale vector $\vh$ that adapts the learning rate.
Specifically, we set the variance as $\vv = 1/(\lambda (\vh + \delta))$ where $\delta$ is the weight decay and $\vh = \nabla^2\ell(\vparam)$ is the diagonal Hessian.
Therefore, $\vv$ can be obtained for free by estimating gradients at a perturbed $\vparam \sim \gauss(\vm,\text{diag}(\vv))$ 
to estimate the expectation in \cref{eq:AdamWvsIVON} (RHS) and using the reparametrization trick to get $\vh$.

In practice, even using one Monte-Carlo sample $\vparam \sim \gauss(\vm,\text{diag}(\vv))$ performs well and incurs almost no overhead (see~\cref{sec:speed} for a benchmark), but more samples can be taken to improve performance. 
Similar perturbed training has also been shown to be useful for finetuning LLMs~\citep{liu2023pac, zhelnin2024gift,li2024flat}.
The hyperparameter $\lambda$ can be seen as an effective training data size, where 
$\lambda=N$ targets a generalized posterior for $N$ data points~\citep{Ze88}.
However, $\lambda$ can also be adapted: for example, $\lambda > N$ targets a ``colder'' posterior~\citep{zhang2006varepsilon,DBLP:conf/alt/Grunwald12} which can
stabilize training~\citep{shen2024variational}.
Conversely, decreasing $\lambda$ during inference can promote more diverse LLM outputs, which can be helpful in scenarios such as combining the outputs of multiple models sampled from IVON during generation~\citep{daheim2025uncertaintyaware}.
For further details, we refer to~\citet{shen2024variational}.
Overall, IVON is an easy-to-use alternative to existing Bayesian approaches that require additional overheads due to post-processing, additional passes through the data, 
and cumbersome implementation changes.

\section{Uncertainty-Guided Pruning (UGP)}
\label{sec:ugp}
To further improve performance, we propose a novel pruning method for IVON-LoRA.
Parameter pruning is usually motivated by computational efficiency, because it can reduce the size of neural networks 
by setting a subset of network parameters to zero.
This can be done either before, during, or after training. 
Intuitively, we might want to set parameters to zero which do not influence the model.
Previously, the posterior variance has been shown to be useful for identifying such unimportant parameters~\citep{Gr11,blundell2015weight,dhahri2024shaving}.
We use it here for an uncertainty-guided pruning of LoRA parameters.

In IVON, high parameter uncertainty is encouraged by using the relative entropy term in the variational objective in \Cref{eq:AdamWvsIVON}.
This should aid in the discovery of prunable LoRA parameters because intuitively such parameters should be the ones that have high uncertainty.
In fact, Hessians have long been used for optimal pruning~\citep{NIPS1989_6c9882bb} and the $\vh$ vector in IVON can be seen as an online estimate of the diagonal Hessian (hence the name Online Newton).
Recently, IVON's posterior variance has also been used for budget allocation in AdaLoRA~\citep{zhang2023adaptive} according to the signal-to-noise ratio $|\param_i|/\sqrt{v_i}$~\citep{chen2024bayesian}, where $\param_i$ and $v_i$ are the $i$-th entry of $\vm$ and $\vv$, respectively.

We take a similar approach and prune parameters $\param_i$ with the largest posterior variance $v_i$.
This prunes parameters with the highest parameter uncertainty.
In practice, we find that pruning a small fraction of parameters with the highest variance in each weight matrix leads to visible improvements in calibration, 
while maintaining or even improving accuracy.
Hence, in our experiments, we set the pruning ratio to 10\% and apply it to the LoRA adapters after training by default.

\section{Experiments}
\label{sec:experiments}

In this section, we evaluate IVON-LoRA on various tasks and datasets.
In~\cref{sec:generalization}, we assess its performance on reasoning and language understanding datasets.
In~\cref{sec:lambda}, we conduct an ablation study on the choice of $\lambda$ at test time, which can bring extra improvements to our method.
In~\cref{sec:pruning}, we investigate the effectiveness of the Uncertainty-guided Pruning (UGP) method.
Finally, in~\cref{sec:speed}, we evaluate the training speed and computational overhead of IVON-LoRA.

\definecolor{verylightgray}{gray}{0.9}
\definecolor{vlg}{gray}{0.6}
\newcommand{\ptm}{\phantom{$_0$}}
\newcommand{\tab}{\hspace{0.3cm}}
\newcommand{\white}{\cellcolor{white}}
\begin{table}[t]
    \setlength{\tabcolsep}{4pt}
    \caption{
        Comparison of methods applied to finetuning/finetuned Llama-3.2-3B model across commonsense reasoning datasets, with subscripts indicating standard error of the mean across 5 runs.
        We show the relative metric improvements achieved over AdamW in \textcolor{blue}{blue}.
    }
    \label{tab:performance}
    \centering
    \small
    \resizebox{\textwidth}{!}{%
    \begin{tabular}{llrrrrrrr}
        \toprule
        \textbf{Metrics} & \textbf{Methods} & \textbf{WG-S}             & \textbf{ARC-C}                & \textbf{ARC-E}                & \textbf{WG-M}                 & \textbf{OBQA}                 & \textbf{BoolQ}                & \textbf{Avg.} \\ 
        \midrule
        \multirow{3}{*}{\textbf{ACC} $\uparrow$} 
        & AdamW                             & 70.3$_{0.7}$              & 72.5$_{0.6}$                  & 88.4$_{0.3}$                  & 77.7$_{0.3}$                  & 81.6$_{0.4}$                  & 86.6$_{0.2}$                  & 79.5 \\
        & BLoB@mean                         & 72.2$_{0.2}$              & 75.7$_{0.7}$                  & 89.3$_{0.2}$                  & 76.5$_{0.4}$                  & 82.3$_{0.3}$                  & 84.9$_{0.2}$                  & {\scriptsize \textcolor{blue}{(+0.6)}} 80.1 \\
        \rowcolor{verylightgray}
\white  & IVON-LoRA@mean                    & 73.2$_{0.5}$              & 75.8$_{0.8}$                  & 89.1$_{0.3}$                  & 77.9$_{0.3}$                  & 82.7$_{0.3}$                  & 85.9$_{0.1}$                  & {\scriptsize \textcolor{blue}{(+1.3)}} 80.8 \\
        \midrule
        \multirow{3}{*}{\makecell{\textbf{ECE} ($\times 100$)} $\downarrow$} 
        & AdamW                             & 28.8$_{0.6}$              & 25.4$_{0.6}$                  & 10.5$_{0.4}$                  & 20.9$_{0.4}$                  & 16.3$_{0.5}$                  & 10.2$_{0.2}$                  & 18.7 \\
        & BLoB@mean                         & 20.9$_{0.3}$              & 17.0$_{0.8}$                  & 6.5$_{0.1}$                   & 11.5$_{0.4}$                  & 7.3$_{0.3}$                   & 2.1$_{0.3}$                   & {\scriptsize \textcolor{blue}{(-7.8)}} 10.9 \\
        \rowcolor{verylightgray}
\white  & IVON-LoRA@mean                    & 23.6$_{0.4}$              & 19.4$_{0.9}$                  & 7.5$_{0.2}$                   & 16.6$_{0.4}$                  & 9.8$_{0.2}$                   & 2.7$_{0.1}$                   & {\scriptsize \textcolor{blue}{(-5.4)}} 13.3 \\
        \midrule
        \multirow{3}{*}{\textbf{NLL} $\downarrow$} 
        & AdamW                             & 3.27$_{0.12}$             & 2.48$_{0.11}$                 & 1.03$_{0.05}$                 & 1.75$_{0.07}$                 & 1.32$_{0.05}$                 & 0.59$_{0.01}$                 & 1.74 \\
        & BLoB@mean                         & 1.02$_{0.01}$             & 1.00$_{0.02}$                 & 0.39$_{0.01}$                 & 0.59$_{0.01}$                 & 0.53$_{0.01}$                 & 0.35$_{0.01}$                 & {\scriptsize \textcolor{blue}{(-1.09)}} 0.65 \\
        \rowcolor{verylightgray}
\white  & IVON-LoRA@mean                    & 1.43$_{0.03}$             & 1.16$_{0.03}$                 & 0.44$_{0.01}$                 & 0.86$_{0.02}$                 & 0.58$_{0.01}$                 & 0.33$_{0.00}$                 & {\scriptsize \textcolor{blue}{(-0.94)}} 0.80 \\
        \bottomrule
    \end{tabular}
    }
\end{table}

\begin{table}[t]
    \setlength{\tabcolsep}{4pt}
    \caption{
        Comparison of methods applied to finetuning/finetuned Llama-3.2-3B model across commonsense reasoning datasets.
        Different from Table~\ref{tab:performance}, methods use a Bayesian approach at test time (model linearization for LA, posterior sampling for BLoB and IVON-LoRA).
        Subscripts indicating standard error of the mean across 5 runs.
        We show the relative metric changes over AdamW in parentheses, with improvements in \textcolor{blue}{blue} and degradation in \textcolor{red}{red}.
    }
    \label{tab:performance_bayesian}
    \centering
    \small
    \resizebox{\textwidth}{!}{%
    \begin{tabular}{llrrrrrrr}
        \toprule
        \textbf{Metrics} & \textbf{Methods} & \textbf{WG-S}             & \textbf{ARC-C}                & \textbf{ARC-E}                & \textbf{WG-M}                 & \textbf{OBQA}                 & \textbf{BoolQ}                & \textbf{Avg.} \\ 
        \midrule
        \multirow{5}{*}{\textbf{ACC} $\uparrow$} 
        & AdamW                             & 70.3$_{0.7}$              & 72.5$_{0.6}$                  & 88.4$_{0.3}$                  & 77.7$_{0.3}$                  & 81.6$_{0.4}$                  & 86.6$_{0.2}$                  & 79.5 \\
        & \tab $+$ LA (KFAC)                & 70.4$_{0.6}$              & 71.5$_{1.1}$                  & 87.9$_{0.7}$                  & 77.4$_{0.4}$                  & 81.4$_{0.3}$                  & 86.7$_{0.2}$                  & {\scriptsize \textcolor{red}{(-0.3)}} 79.2 \\
        & \tab $+$ LA (diag)                & 70.4$_{0.5}$              & 61.8$_{0.3}$                  & 80.5$_{0.3}$                  & 77.5$_{0.4}$                  & 81.4$_{0.3}$                  & 86.7$_{0.2}$                  & {\scriptsize \textcolor{red}{(-3.1)}} 76.4 \\
        & BLoB                              & 68.4$_{0.5}$              & 73.7$_{0.5}$                  & 88.6$_{0.3}$                  & 73.1$_{0.3}$                  & 81.8$_{0.3}$                  & 84.6$_{0.2}$                  & {\scriptsize \textcolor{red}{(-1.1)}} 78.4 \\
        \rowcolor{verylightgray}
\white  & IVON-LoRA                         & 72.5$_{0.4}$              & 75.7$_{0.5}$                  & 89.4$_{0.2}$                  & 77.8$_{0.4}$                  & 83.1$_{0.2}$                  & 85.8$_{0.2}$                  & {\scriptsize \textcolor{blue}{(+1.2)}} 80.7 \\
        \midrule
        \multirow{5}{*}{\makecell{\textbf{ECE} \\ ($\times 100$)} $\downarrow$} 
        & AdamW                             & 28.8$_{0.6}$              & 25.4$_{0.6}$                  & 10.5$_{0.4}$                  & 20.9$_{0.4}$                  & 16.3$_{0.5}$                  & 10.2$_{0.2}$                  & 18.7 \\
        & \tab $+$ LA (KFAC)                & 8.3$_{1.2}$               & 11.4$_{1.3}$                  & 15.1$_{1.2}$                  & 5.7$_{0.2}$                   & 4.6$_{0.8}$                   & 4.4$_{0.1}$                   & {\scriptsize \textcolor{blue}{(-10.5)}} \ptm{}8.2  \\
        & \tab $+$ LA (diag)                & 17.1$_{0.7}$              & 18.9$_{0.5}$                  & 34.5$_{0.5}$                  & 14.8$_{0.6}$                  & 17.1$_{1.0}$                  & 17.1$_{0.4}$                  & {\scriptsize \textcolor{red}{(+1.2)}} 19.9 \\
        & BLoB                              & 11.8$_{0.4}$              & 5.9$_{0.8}$                   & 4.1$_{0.2}$                   & 3.2$_{0.1}$                   & 3.8$_{0.5}$                   & 3.8$_{0.2}$                   & {\scriptsize \textcolor{blue}{(-13.3)}} \ptm{}5.4  \\
        \rowcolor{verylightgray}
\white  & IVON-LoRA                         & 19.1$_{0.3}$              & 14.1$_{0.4}$                  & 4.6$_{0.1}$                   & 13.9$_{0.5}$                  & 7.6$_{0.3}$                   & 1.2$_{0.2}$                   & {\scriptsize \textcolor{blue}{(-8.6)}} 10.1 \\
        \midrule
        \multirow{5}{*}{\textbf{NLL} $\downarrow$} 
        & AdamW                             & 3.27$_{0.12}$             & 2.48$_{0.11}$                 & 1.03$_{0.05}$                 & 1.75$_{0.07}$                 & 1.32$_{0.05}$                 & 0.59$_{0.01}$                 & 1.74 \\
        & \tab $+$ LA (KFAC)                & 0.60$_{0.00}$             & 0.79$_{0.01}$                 & 0.45$_{0.01}$                 & 0.52$_{0.01}$                 & 0.58$_{0.01}$                 & 0.38$_{0.00}$                 & {\scriptsize \textcolor{blue}{(-1.19)}} 0.55 \\
        & \tab $+$ LA (diag)                & 0.66$_{0.00}$             & 1.02$_{0.01}$                 & 0.85$_{0.01}$                 & 0.55$_{0.00}$                 & 0.61$_{0.01}$                 & 0.43$_{0.01}$                 & {\scriptsize \textcolor{blue}{(-1.05)}} 0.69 \\
        & BLoB                              & 0.67$_{0.00}$             & 0.73$_{0.01}$                 & 0.34$_{0.01}$                 & 0.55$_{0.01}$                 & 0.50$_{0.00}$                 & 0.36$_{0.01}$                 & {\scriptsize \textcolor{blue}{(-1.22)}} 0.52 \\
        \rowcolor{verylightgray}
\white  & IVON-LoRA                         & 1.04$_{0.01}$             & 0.92$_{0.02}$                 & 0.36$_{0.01}$                 & 0.73$_{0.01}$                 & 0.52$_{0.01}$                 & 0.33$_{0.00}$                 & {\scriptsize \textcolor{blue}{(-1.09)}} 0.65 \\
        \bottomrule
    \end{tabular}
    }
\end{table}

\subsection{IVON-LoRA Improves Accuracy, Generalization and Calibration}
\label{sec:generalization}

\subsubsection{Results on Commonsense Reasoning Datasets}
\label{sec:commonsense}

Following the settings in~\citet{yang2023bayesian} and~\citet{wang2024blob}, we begin with evaluating the performance of our 
method on commonsense reasoning tasks.
We use IVON-LoRA to finetune Llama-3.2-3B~\citep{grattafiori2024llama} on six commonsense reasoning datasets, including WinoGrande-S
(WG-S), WinoGrande-M (WG-M)~\citep{sakaguchi2021winogrande}, ARC-Challenge (ARC-C), ARC-Easy (ARC-E)~\citep{clark2018think}, 
OpenBookQA (OBQA)~\citep{mihaylov2018can}, and BoolQ~\citep{clark2019boolq}.
We then measure accuracy and Expected Calibration Error (ECE) on the validation splits and use Negative Log-Likelihood (NLL) as an additional metric for calibration since ECE 
may be unreliable when annotators disagree~\citep{baan-etal-2022-stop}.
We compare our results to baselines including non-Bayesian LoRA finetuning with AdamW, Laplace-LoRA~\citep{yang2023bayesian} and BLoB~\citep{wang2024blob}.
For BLoB and IVON-LoRA, we report the results acquired by either using the mean of the posterior $\vm$ (indicated by the suffix ``@mean'') or by using an averaged prediction over 10 samples from the posterior.
Please refer to~\cref{app:commonsense} for details on the experimental setup.

Results are shown in \cref{tab:performance,tab:performance_bayesian}.
As shown in~\cref{tab:performance}, IVON-LoRA@mean outperforms baseline methods on most datasets in terms of accuracy, often by a large margin.
IVON-LoRA@mean on average improves accuracy by 1.3\% over AdamW and 0.7\% over BLoB@mean.
Our method also exhibits significantly improved calibration compared to AdamW baseline.
Notably, these improvements are achieved with no test-time overhead at all, since only the mean of the posterior is used for prediction.
Next, as shown in~\cref{tab:performance_bayesian}, IVON-LoRA performs significantly better than baseline methods in a more Bayesian setting.
IVON-LoRA with posterior sampling still maintains an 1.2\% accuracy gain over AdamW, while both BLoB (with posterior sampling) and Laplace-LoRA \textit{degrade} accuracy.
With posterior sampling, IVON-LoRA further reduces average ECE from 13.3 to 10.1 and NLL from 0.80 to 0.65.
Notably, this is achieved with neither significant drop in accuracy (as in BLoB) nor a more representative KFAC Hessian or an additional pass through the data to compute that Hessian (as in Laplace-LoRA).

\subsubsection{IVON-LoRA for LLM Reasoning}
\begin{table*}[t]
    \centering
    \resizebox{.95\textwidth}{!}{\begin{tabular}{lccccc}
        \toprule
        \textbf{Method}       & \bf GSM8k & \multicolumn{4}{c}{\bf Conala} \\
        & Accuracy & CodeBLEU$\uparrow$ & Syntax$\uparrow$ & Data Flow$\uparrow$ & Code-BertScore$\uparrow$ \\
        \midrule
        AdamW & 66.87 & \bf 28.73 & 40.55 & 40.97 & \bf 87.89 \\
        \midrule
        \rowcolor{verylightgray} IVON-LoRA@mean & 68.31 & 28.50 & \bf 41.23 & 40.80 & 87.34 \\
        \rowcolor{verylightgray} IVON-LoRA & \bf 68.69 & 28.45 & 41.19 & \bf 41.05 & 87.65 \\
        \bottomrule
    \end{tabular}}
    \caption{IVON-LoRA can improve the performance of LLMs for reasoning, here on math word problem solving and code generation with uncertainty-aware MBR with 32 outputs.
    }
    \label{tab:reasoning}
\end{table*}

Next, we evaluate IVON-LoRA on LLM reasoning.
We train Qwen-2.5-3B~\citep{qwen2025qwen25technicalreport} on the GSM8k benchmark for math word problem solving~\citep{cobbe2021gsm8k} 
and the Conala benchmark for Python code generation~\citep{yin2018learning}.
We compare IVON-LoRA evaluated at the learned mean and using the posterior against training with LoRA and AdamW.
For using the posterior we use sequence-level uncertainty-aware Minimum Bayes Risk (MBR) decoding~\citep[Eq. 9]{daheim2025uncertaintyaware}.
That is, we first sample multiple outputs for each model sampled from the IVON-LoRA posterior. 
Then we use a utility function to compare each pair of outputs in the resulting n-best list of size $n$.
For GSM8k we use a 0-1-loss and perform majority voting on the final numerical solution.
For Conala we use CodeBertScore~\citep{zhou-etal-2023-codebertscore}.

For IVON-LoRA@mean and AdamW we sample 32 outputs.
When using the posterior, we sample 4 models and then 8 outputs per model to match compute budget.
For details on the experimental setup, please refer to~\cref{app:gen}.
Results are shown in~\cref{tab:reasoning}.
We find that both IVON-LoRA@mean and IVON-LoRA at the posterior provide strong improvements over AdamW for GSM8k.
For Conala, we find improvements especially in syntax and data flow matching when compared to a reference solution, as well as comparable CodeBertScore and CodeBLEU~\citep{ren2020codebleu}.

\subsubsection{IVON-LoRA Enables Test-Time Compute Scaling}
\begin{figure}[t!]
    \centering
\begin{subfigure}{.48\textwidth}
    \center
    \includegraphics[width=\linewidth]{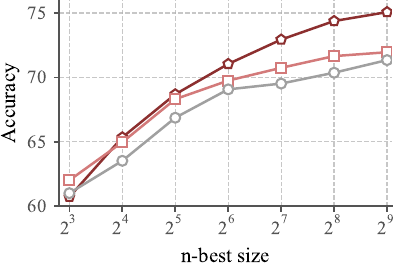}
    \includegraphics[width=\linewidth]{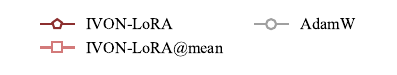}
    \caption{IVON vs. AdamW}
    \label{fig:mbr1}
\end{subfigure}\hfill%
\begin{subfigure}{.48\textwidth}
    \center
    \includegraphics[width=\linewidth]{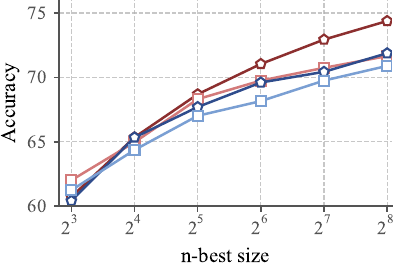}
    \includegraphics[width=\linewidth]{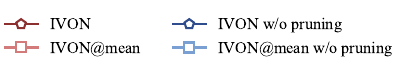}
    \caption{Effect of pruning}
    \label{fig:mbr2}
\end{subfigure}
\caption{Improvements obtained with IVON-LoRA on GSM8k increase with n-best-size. For smaller $n$ IVON-LoRA@mean can be most efficient (a).
Furthermore, we show that pruning is essential for this, because high-uncertainty parameters are not included when sampling models (b).
}
\end{figure}

\newcolumntype{Y}{>{\centering\arraybackslash}X} 
\begin{table}[t]
    \centering
    \renewcommand{\arraystretch}{1.1}
    \small
  
    \resizebox{\textwidth}{!}{%
    \begin{tabular}{llccccccccc}
      \toprule
    \multirow{4}{*}{\textbf{Metric}} &
    \multirow{4}{*}{\textbf{Method}} &
    \multicolumn{9}{c}{\textbf{Datasets}} \\
    \cmidrule(lr){3-11}
    & & 
    \multicolumn{1}{c}{\textit{In-Dist.}} &
    \multicolumn{2}{c}{\textit{Smaller Dist.~Shift}} &
    \multicolumn{6}{c}{\textit{Larger Dist.~Shift}} \\
    \cmidrule(lr){3-3}\cmidrule(lr){4-5}\cmidrule(lr){6-11}
    & &
    \textbf{OBQA} & \textbf{ARC-C} & \textbf{ARC-E} & \textbf{Chem} & \textbf{Phy} & \textbf{Bio} & \textbf{CS} & \textbf{Math} & \textbf{Avg.} \\
    \midrule
    \multirow{8}{*}{\textbf{ACC} $\uparrow$}
        & AdamW\textsuperscript                 & 81.6$_{0.4}$              & 70.1$_{0.4}$              & 79.4$_{0.4}$              & 40.4$_{2.2}$              & 37.3$_{1.8}$              & 55.0$_{1.6}$              & \textbf{42.8}$_{1.9}$     & \underline{35.2$_{1.2}$}  & 42.1\\
        & \tab $+$ LA (KFAC)                    & 81.4$_{0.3}$              & 71.0$_{0.7}$              & 79.7$_{0.2}$              & \textbf{42.6}$_{1.9}$     & 38.4$_{1.9}$              & 55.0$_{0.9}$              & 39.6$_{1.4}$              & 35.0$_{0.8}$              & 42.1\\
        & \tab $+$ LA (diag)                    & 81.4$_{0.3}$              & 70.4$_{0.6}$              & 79.0$_{0.6}$              & \underline{42.0$_{1.1}$}  & \textbf{40.4}$_{1.4}$     & 54.4$_{1.1}$              & 37.4$_{1.7}$              & \textbf{36.4}$_{0.7}$     & 42.1\\
        & BLoB@mean\textsuperscript             & 82.3$_{0.3}$              & \textbf{72.7}$_{0.5}$     & 80.0$_{0.5}$              & 33.6$_{0.7}$              & 38.4$_{1.2}$              & \underline{58.8$_{0.7}$}  & 42.4$_{0.7}$              & 33.0$_{0.8}$              & 41.2 \\
        & BLoB                                  & 81.8$_{0.3}$              & 71.5$_{0.8}$              & 78.3$_{0.3}$              & 34.4$_{0.9}$              & 34.9$_{1.0}$              & 56.5$_{0.3}$              & 40.8$_{1.3}$              & 31.4$_{1.3}$              & 39.6\\
        \rowcolor{verylightgray}
\white  & IVON-LoRA@mean\textsuperscript        & \underline{82.7$_{0.3}$}  & 71.2$_{0.4}$              & \textbf{81.2}$_{0.6}$     & 38.4$_{0.9}$              & 39.0$_{1.7}$              & 58.2$_{0.8}$              & \underline{42.6$_{1.4}$}  & 33.8$_{1.8}$              & 42.4\\
        \rowcolor{verylightgray}
\white  & IVON-LoRA                             & \textbf{83.1}$_{0.2}$     & \underline{71.9$_{0.3}$}  & \underline{81.1$_{0.5}$}  & 36.4$_{1.2}$              & \underline{39.4$_{1.7}$}  & \textbf{58.9}$_{1.2}$     & 42.4$_{1.1}$              & 32.0$_{0.8}$              & 41.8\\

    \midrule
    \multirow{8}{*}{\makecell{\textbf{ECE} \\ ($\times 100$)} $\downarrow$}
        & AdamW\textsuperscript                 & 16.3$_{0.5}$              & 26.6$_{0.3}$              & 18.0$_{0.4}$              & 32.9$_{2.1}$              & 38.1$_{1.4}$              & 32.1$_{2.2}$              & 32.9$_{1.0}$              & 35.2$_{1.0}$              & 34.2\\
        & \tab $+$ LA (KFAC)                    & \underline{4.6$_{0.8}$}   & \textbf{9.1}$_{0.4}$      & \textbf{4.5}$_{0.5}$      & \underline{10.2$_{1.9}$}  & 13.7$_{1.9}$              & \textbf{7.6}$_{1.0}$      & 13.7$_{1.4}$              & 14.1$_{0.5}$              & 11.9\\
        & \tab $+$ LA (diag)                    & 17.1$_{1.0}$              & 9.4$_{0.7}$               & 14.0$_{0.9}$              & \textbf{5.5}$_{1.0}$      & \textbf{7.6}$_{1.5}$      & 9.3$_{1.0}$               & 11.7$_{1.1}$              & \textbf{8.0}$_{1.2}$      & 8.4\\
        & BLoB@mean\textsuperscript             & 7.3$_{0.3}$               & 14.9$_{0.7}$              & 10.8$_{0.4}$              & 20.4$_{1.0}$              & 16.0$_{0.8}$              & 13.4$_{1.0}$              & 14.1$_{0.4}$              & 16.9$_{2.1}$              & 16.2 \\
        & BLoB                                  & \textbf{3.8}$_{0.5}$      & \underline{9.3$_{0.6}$}   & \underline{4.9$_{0.4}$}   & 14.5$_{1.0}$              & 13.7$_{1.5}$              & \underline{8.7$_{0.4}$}   & \underline{11.1$_{1.6}$}  & \underline{12.3$_{1.2}$}  & 12.0\\
        \rowcolor{verylightgray}
\white  & IVON-LoRA@mean\textsuperscript        & 9.8$_{0.2}$               & 18.1$_{0.4}$              & 10.5$_{0.7}$              & 13.4$_{0.7}$              & 14.0$_{1.5}$              & 11.4$_{0.2}$              & 12.1$_{1.6}$              & 14.4$_{1.5}$              & 13.1\\
    \rowcolor{verylightgray}
\white  & IVON-LoRA                             & 7.6$_{0.3}$               & 14.9$_{0.4}$              & 8.8$_{0.5}$               & 14.2$_{1.1}$              & \underline{12.4$_{1.7}$}  & 9.8$_{0.8}$               & \textbf{9.5}$_{1.8}$      & 14.0$_{1.5}$              & 12.0\\

    \midrule
    \multirow{8}{*}{\textbf{NLL} $\downarrow$}
        & AdamW\textsuperscript                 & 1.32$_{0.05}$             & 2.28$_{0.06}$             & 1.56$_{0.04}$             & 2.16$_{0.10}$             & 2.29$_{0.07}$             & 2.14$_{0.14}$             & 2.07$_{0.04}$             & 2.13$_{0.07}$             & 2.16\\
        & \tab $+$ LA (KFAC)                    & 0.58$_{0.01}$             & 0.99$_{0.03}$             & 0.68$_{0.02}$             & 1.39$_{0.03}$             & 1.46$_{0.03}$             & 1.17$_{0.04}$             & 1.42$_{0.03}$             & 1.44$_{0.01}$             & 1.37\\
        & \tab $+$ LA (diag)                    & 0.61$_{0.01}$             & \underline{0.81$_{0.01}$} & 0.65$_{0.01}$             & \underline{1.32$_{0.01}$} & 1.36$_{0.02}$             & 1.10$_{0.02}$             & 1.33$_{0.01}$             & 1.37$_{0.01}$             & 1.30\\
        & BLoB@mean\textsuperscript             & 0.53$_{0.01}$             & 0.95$_{0.02}$             & 0.65$_{0.01}$             & 1.42$_{0.01}$             & 1.41$_{0.01}$             & \textbf{1.05}$_{0.02}$    & 1.30$_{0.01}$             & 1.45$_{0.02}$             & 1.33 \\
        & BLoB                                  & \textbf{0.50}$_{0.00}$    & \textbf{0.79}$_{0.01}$    & \textbf{0.56}$_{0.01}$    & \textbf{1.31}$_{0.01}$    & 1.38$_{0.01}$             & 1.07$_{0.01}$             & \textbf{1.29}$_{0.01}$    & 1.38$_{0.01}$             & 1.29\\
        \rowcolor{verylightgray}
\white  & IVON-LoRA@mean\textsuperscript        & 0.58$_{0.01}$             & 1.05$_{0.01}$             & 0.66$_{0.01}$             & 1.38$_{0.02}$             & \underline{1.34$_{0.02}$} & 1.08$_{0.02}$             & 1.30$_{0.02}$             & \textbf{1.35}$_{0.00}$    & 1.29\\
        \rowcolor{verylightgray}
\white  & IVON-LoRA                             & \underline{0.52$_{0.01}$} & 0.94$_{0.01}$             & \underline{0.61$_{0.01}$} & 1.35$_{0.01}$             & \textbf{1.32}$_{0.01}$    & \underline{1.06$_{0.02}$} & \textbf{1.29}$_{0.01}$    & \textbf{1.35}$_{0.01}$    & 1.27\\
    \bottomrule
    \vspace{0.1cm}
  \end{tabular}
    }

  \caption{
    Comparison of different methods on in- and out-of-distribution scenarios.
    We use the LoRA adapters trained on OBQA and evaluate their performance on different levels of distribution shift.
    We find that IVON-LoRA can cope well with such distribution shifts and maintains better or similar accuracy as well as better calibration than LoRA with AdamW.
}
    \label{tab:ood}
\end{table}

Next, we repeat the experiment on GSM8k but scale the size of the n-best list up to 512 output samples in total.
For IVON-LoRA@mean and AdamW we just sample the outputs from one model.
For IVON-LoRA with posterior we sample 8 outputs each from (n-best-list-size$/8$) models, i.e. we use 64 model samples from the posterior for the $n=512$ and sample 8 outputs from each model.
\cref{fig:mbr1} shows that IVON@mean outperforms AdamW especially for smaller $n$.
As $n$ and the number of models grow the benefit of the posterior become more and more apparent with an accuracy improvement of $3.7\%$ for $n=512$.
In~\cref{fig:mbr2} we further show that pruning is essential for this, with large improvements for both IVON-LoRA@mean and IVON-LoRA when it is used.
Intuitively, this might be, because sampling high-uncertainty parameters can lead to destructive behavior in the model.
Altogether, IVON-LoRA provides a strong and easy-to-use method for test-time compute scaling.

\subsubsection{Results for Out-of-Distribution Settings}
\label{sec:ood}

Here, we evaluate the performance of IVON-LoRA under out-of-distribution settings.
Following~\citet{yang2023bayesian} and~\citet{wang2024blob}, we evaluate the performance of LoRA adapters trained on OBQA, as in~\cref{sec:commonsense},
on datasets with different levels of distribution shifts.
Specifically, we use ARC-E and ARC-C for smaller and the college-level chemistry, physics, biology, computer science, and math subsets from the MMLU benchmark~\cite{hendrycks2021measuring} for larger distribution shifts.
Results are shown in \cref{tab:ood}.
IVON-LoRA can also improve accuracy under mild distribution shifts (indicated by the 1.8\% improvements on ARC-C and 1.7\% on ARC-E) 
and still achieves good calibration.
Under severe distribution shifts IVON-LoRA still maintains similarly good calibration as Laplace-LoRA and BLoB, while standard LoRA finetuning with AdamW exhibits significant overconfidence.

\begin{table*}[t]
    \caption{Performance comparison on the test sets of GLUE benchmark using DeBERTa-v3-base as the base model. Results are averaged over 5 runs with different random seeds.}
    \label{tab:glue}
    \centering
    \begin{tabular}{lcccccccccc}
        \toprule
        \textbf{Method} & \textbf{CoLA} & \textbf{MRPC} & \textbf{RTE} & \textbf{STS-B} & \textbf{QQP} & \textbf{QNLI} & \textbf{SST-2} & \textbf{Avg.} \\
        \midrule
        Full-FT         & 68.0$_{0.5}$  & 90.1$_{0.5}$  & 79.4$_{1.4}$  & 90.9$_{0.1}$  & 90.6$_{0.0}$  & 94.0$_{0.1}$  & 95.5$_{0.1}$  & 86.9  \\
        \midrule
        Vanilla LoRA    & 68.6$_{0.9}$  & 88.9$_{0.4}$  & 84.0$_{1.1}$  & 90.9$_{0.1}$  & 91.1$_{0.0}$  & 94.3$_{0.1}$  & 95.3$_{0.1}$  & 87.6  \\
        IVON-LoRA       & 68.6$_{0.2}$  & 90.6$_{0.3}$  & 86.2$_{0.5}$  & 91.1$_{0.1}$  & 90.7$_{0.0}$  & 94.1$_{0.1}$  & 95.4$_{0.1}$  & 88.1  \\
        \bottomrule
        \end{tabular}
\end{table*}

\definecolor{verylightgray}{gray}{0.9}
\begin{table}[t]
    \setlength{\tabcolsep}{4pt}
    \caption{
        Evaluating OBQA-finetuned Llama-3.2-3B model on out-of-distribution MMLU subsets. The setting is the same as in~\cref{sec:ood}, except varying effective sample size or inverse temperature of the posterior.
        Sampling with lower temperature can improve accuracy at similar calibration.
    }
    \label{tab:ablation-ess}
    \centering
    \small
    \begin{tabular}{llcccccc}
        \toprule
        \textbf{Metrics} & \textbf{Methods}     & \textbf{Chem}             & \textbf{Phy}                & \textbf{Bio}                & \textbf{CS}                 & \textbf{Math}                 & \textbf{Avg.} \\ 
        \midrule
        \multirow{5}{*}{\textbf{ACC} $\uparrow$} 
        & IVON-LoRA ($\tau=1$)                  & 36.4$_{1.2}$  & 39.4$_{1.7}$  & 58.9$_{1.2}$  & 42.4$_{1.1}$  & 32.0$_{0.8}$  & 41.8 \\
        & \tab $\tau=2$                         & 37.0$_{1.5}$  & 39.0$_{1.8}$  & 59.6$_{1.2}$  & 43.0$_{0.9}$  & 32.8$_{1.9}$  & 42.3 \\
        & \tab $\tau=5$                         & 38.2$_{1.2}$  & 40.4$_{1.8}$  & 59.9$_{1.1}$  & 43.4$_{0.5}$  & 32.8$_{2.0}$  & \textbf{42.9} \\
        & \tab $\tau=10$                        & 39.0$_{0.9}$  & 40.0$_{1.4}$  & 59.9$_{1.3}$  & 42.6$_{0.5}$  & 32.4$_{1.1}$  & \underline{42.8} \\
        & IVON-LoRA@mean ($\tau \to \infty$)    & 38.4$_{0.9}$  & 39.0$_{1.7}$  & 58.2$_{0.8}$  & 42.6$_{1.4}$  & 33.8$_{1.8}$  & 42.4 \\
        \midrule
        \multirow{5}{*}{\makecell{\textbf{ECE} \\ ($\times 100$)} $\downarrow$} 
        & IVON-LoRA ($\tau=1$)                  & 14.2$_{1.1}$  & 12.4$_{1.7}$  & 9.8$_{0.8}$   & 9.5$_{1.8}$   & 14.0$_{1.5}$  & \underline{12.0} \\
        & \tab $\tau=2$                         & 15.5$_{1.5}$  & 13.1$_{1.1}$  & 12.8$_{0.9}$  & 10.3$_{0.6}$  & 15.2$_{0.9}$  & 13.4 \\
        & \tab $\tau=5$                         & 13.3$_{1.2}$  & 10.4$_{1.4}$  & 12.0$_{0.6}$  & 10.7$_{0.8}$  & 13.9$_{1.4}$  & 12.1 \\
        & \tab $\tau=10$                        & 13.0$_{0.8}$  & 11.1$_{1.4}$  & 11.7$_{0.8}$  & 10.2$_{0.7}$  & 13.3$_{0.8}$  & \textbf{11.9} \\
        & IVON-LoRA@mean ($\tau \to \infty$)    & 13.4$_{0.7}$  & 14.0$_{1.5}$  & 11.4$_{0.2}$  & 12.1$_{1.6}$  & 14.4$_{1.5}$  & 13.1 \\
        \midrule
        \multirow{5}{*}{\textbf{NLL} $\downarrow$}
        & IVON-LoRA ($\tau=1$)                  & 1.35$_{0.01}$         & 1.32$_{0.01}$         & 1.06$_{0.02}$         & 1.29$_{0.01}$         & 1.35$_{0.01}$         & \textbf{1.27} \\
        & \tab $\tau=2$                         & 1.37$_{0.01}$         & 1.32$_{0.02}$         & 1.07$_{0.02}$         & 1.30$_{0.02}$         & 1.35$_{0.00}$         & \underline{1.28} \\
        & \tab $\tau=5$                         & 1.36$_{0.01}$         & 1.33$_{0.02}$         & 1.07$_{0.02}$         & 1.29$_{0.01}$         & 1.35$_{0.00}$         & \underline{1.28} \\
        & \tab $\tau=10$                        & 1.36$_{0.01}$         & 1.33$_{0.02}$         & 1.07$_{0.02}$         & 1.30$_{0.02}$         & 1.35$_{0.01}$         & \underline{1.28} \\
        & IVON-LoRA@mean ($\tau \to \infty$)    & 1.38$_{0.02}$         & 1.34$_{0.02}$         & 1.08$_{0.02}$         & 1.30$_{0.02}$         & 1.35$_{0.00}$         & 1.29 \\
        \bottomrule
    \end{tabular}
\end{table}

\subsubsection{Results on GLUE}

In this section, we evaluate IVON-LoRA on the GLUE benchmark~\citep{wang2018glue}.
We use DeBERTa-v3-base~\citep{he2021deberta} and compare our method to full-parameter finetuning and LoRA with AdamW.
We only evaluate the performance of IVON-LoRA at the mean of the posterior and do not employ UGP.
Please refer to~\cref{app:glue} for details on the experimental setup.
We present the results in~\cref{tab:glue}. Similar to the results in~\citet{hu2021lora}, we observe that LoRA achieves a higher average score than full-parameter finetuning.
Notably, IVON-LoRA outperforms vanilla LoRA by 0.5 on average.

\begin{figure}[t]
    \centering
    \includegraphics[width=\textwidth]{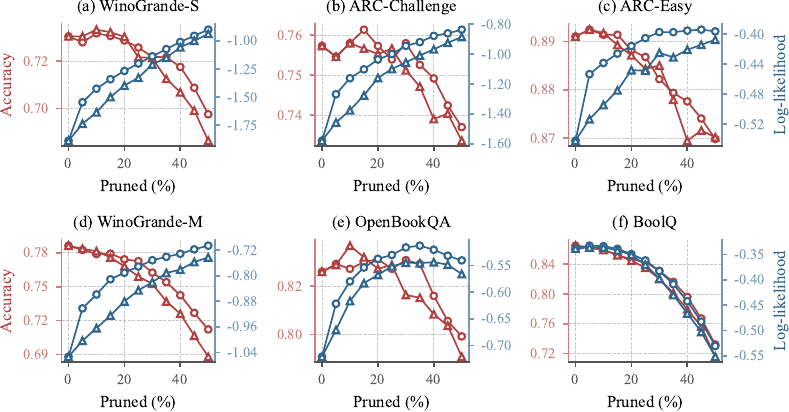}
    \includegraphics[width=\textwidth]{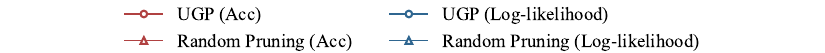}
    \caption{
      Uncertainty-Guided Pruning is essential for improving the performance of IVON-LoRA. 
      We show the accuracy and log-likelihood of IVON-LoRA on commonsense reasoning, with the x-axis indicating the pruning strength.
      We observe that UGP can significantly improve calibration with sometimes even minor improvement in accuracy, while also outperforming random pruning.
    }
    \label{fig:pruning}
  \end{figure}
  \begin{figure}[t]
      \centering
      \includegraphics[width=\textwidth]{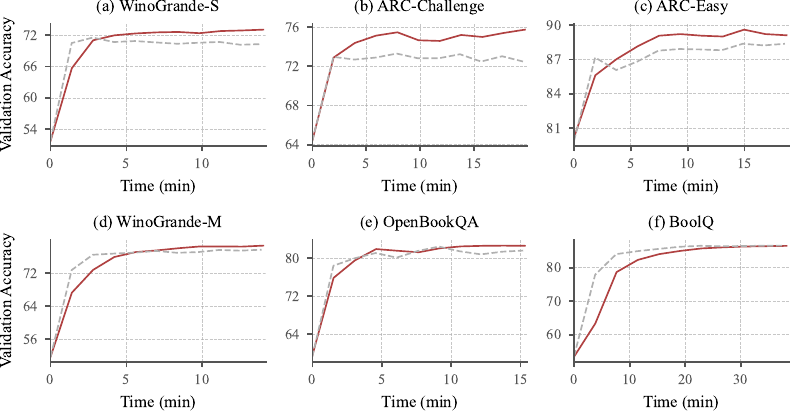}
      \includegraphics[width=\textwidth]{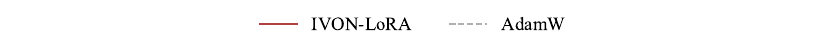}
      \caption{
        The training speeds of IVON and AdamW are similar.
        We plot validation accuracies (without pruning) of the two methods versus time in minutes.
        Results are averaged over 5 runs.
        }
      \label{fig:curve}
    \end{figure}

\subsection{Adjusting Temperature Can Improve Performance}
\label{sec:lambda}
As described in~\cref{sec:method}, the $\lambda$ parameter in IVON-LoRA's posterior variance can be adjusted during test time to improve performance.
In particular, we find that using a moderately larger $\lambda_\text{test}=\tau\lambda$ can further improve generalization with minimal degradation in calibration.
To show this, we evaluate IVON-LoRA on the larger distribution shift setting from Sec.~\ref{sec:ood} with different $\lambda_\text{test}$.
Specifically, we test the performance with $\tau$ set to 2, 5 and 10.
Results are shown in \cref{tab:ablation-ess}.
Notably, setting $\tau=5$ and $\tau=10$ improves the average accuracy of IVON-LoRA by 1.1\% and 1\%, respectively, with virtually no increases in ECE and NLL.

\subsection{UGP Improves Calibration and Accuracy}
\label{sec:pruning}

Here we show that our parameter pruning method is effective by evaluating on the commonsense reasoning datasets from Sec.~\ref{sec:commonsense}.
We test the performance under different pruning strengths ranging from 0\% to 50\% of the parameters, and compare the performance of our method with random pruning.
The results are shown in \cref{fig:pruning}.
First, we observe that pruning a small number of parameters can significantly improve the calibration with little degradation or sometimes even minor improvement in accuracy.
Second, we observe that UGP consistently maintains higher accuracy and log-likelihood across varying pruning levels when compared with random pruning, 
showing that IVON's posterior variance can be a good indicator of parameter importance.
Overall, our results show that UGP can be a simple but effective way to improve the generalization and calibration of IVON-LoRA.

\subsection{Computational Efficiency}
\label{sec:speed}

Finally, we observe that the overhead of IVON-LoRA is negligible compared to AdamW.
We profile our training code on an NVIDIA GeForce RTX 4090 GPU.
In our test run with WinoGrande-S dataset, the forward pass, loss computation, and backward pass of a training step take in total 167.0ms on average.
As for the overhead of IVON, the sampling procedure and the optimization step of each training step take 1.0ms and 0.5ms on average, respectively, 
which is less than 1\% of the per-step running time.
The overall training speed of IVON-LoRA and AdamW are similar as shown in \cref{fig:curve}.

\section{Limitations}
\label{sec:limitations}
A limitation shared with other Bayesian LoRA methods~\citep{yang2023bayesian,onal2024gaussian} is that the learned posterior over the increment low-rank parameters is non-Gaussian because it is a product of two Gaussian random variables. 
If this is a problem, a workaround could be to use a variational low-rank correction to correct the mean and variance of a Laplace approximation of the original model.~\citet{van2024low} propose such a low-rank approach in the context of latent Gaussian models, and adapting these ideas to large language models may represent an interesting direction for future work.

\section{Conclusion}
In this work we introduce IVON-LoRA, a method that improves LoRA by using variational learning.
IVON-LoRA requires minimal change to the training procedure and has minimal overhead over AdamW,  
yet it learns a posterior distribution over parameters instead of a single point estimate.
Despite its simplicity, we show that IVON-LoRA consistently improves the generalization and calibration of LoRA finetuning across tasks, outperforming AdamW and other Bayesian baselines such as Laplace-LoRA and BLoB.
We also investigate its ability to support test-time scaling by posterior sampling and MBR decoding, the effectiveness of uncertainty-guided pruning (UGP), and the impact of adjusting the effective sample size $\lambda$ at test time.
Our results suggest that variational learning with IVON is a simple yet effective way to improve LoRA for large language models.

\begin{ack}
This work is supported by JST CREST Grant Number JPMJCR2112.
This work used computational resources TSUBAME4.0 supercomputer provided by Institute of Science Tokyo through the HPCI System Research Project (Project ID: hp240170).
This research work has been funded by the German Federal Ministry of Education and Research and the Hessian Ministry of Higher Education, Research, Science and the Arts within their joint support of the National Research Center for Applied Cybersecurity ATHENE.
\end{ack}

\small
\bibliographystyle{plainnat}
\bibliography{main}
\normalsize

\newpage

\appendix

\section{Details on Experimental Setup}

\subsection{General Setup}

We utilize the PEFT~\citep{peft} library for LoRA adaptation, and apply LoRA to the query and value weights of the attention layers.
Unlike~\citet{yang2023bayesian}, we do not apply LoRA to the output layer due to numerical instability encountered in some preliminary experiments.
For all experiments with LoRA, we set the rank $r$ to 8, $\alpha$ to 16, and LoRA dropout rate to 0.1, which are the default settings in the PEFT~\citep{peft} library.

\subsection{Commonsense Reasoning}
\label{app:commonsense}

We finetune Llama-3.2-3B~\citep{grattafiori2024llama} on six commonsense reasoning datasets, including WinoGrande-S (WG-S), WinoGrande-M (WG-M)~\citep{sakaguchi2021winogrande}, ARC-Challenge (ARC-C), ARC-Easy (ARC-E)~\citep{clark2018think}, OpenBookQA (OBQA)~\citep{mihaylov2018can}, and BoolQ~\citep{clark2019boolq}.
For better computational efficiency, the base model undergoes int8 quantization, with LoRA weights maintained in 16-bit precision.
Finetuning is performed on a single NVIDIA A100 Tensor Core GPU (with 80 GB memory, for BoolQ dataset due to the memory requirements for longer context) or GeForce RTX 4090 GPU (with 24 GB memory, for other datasets).

To finetune a pretrained language model which predicts the next token in a sequence for solving multiple-choice or true/false questions,
we need to wrap the text and the choice of each question into an instruction using predefined prompt templates.
We then use the pretrained model to predict the next token of the wrapped instruction,
and extract the output logits for the tokens standing for "True"/"False" or "A"/"B"/"C"/"D" choices.
For the prompt templates, we use the same ones as in~\citet{yang2023bayesian}, which are shown in \cref{tab:prompt}.

\begin{table}[h]
    \caption{Prompt templates used for preprocessing different commonsense reasoning datasets.}
    \label{tab:prompt}
    \centering
    \begin{tabular}{lp{0.798\linewidth}}
    \toprule
    \textbf{Dataset} & \textbf{Prompt Template} \\
    \midrule
    BoolQ       & Answer the question with only True or False: \texttt{\{question\}} Context: \texttt{\{passage\}}\\
    \midrule
    OBQA / ARC  & Select one of the choices that answers the following question: \texttt{\{question\}} Choices: \texttt{\{choices\}} Answer:\\
    \midrule
    WG          & Select one of the choices that answers the following question: \texttt{\{sentence\}} Choices: A. \texttt{\{option1\}} B. \texttt{\{option2\}} Answer:\\
    \bottomrule
    \end{tabular}
\end{table}

\subsubsection{Hyperparameters}
\paragraph{General Settings}
For all methods, we set the batch size to 8 and the number of training steps to 5,000.
We use a linear learning rate scheduler with 150 steps of warmup for AdamW and IVON-LoRA and 300 steps for BLoB (which is its default).

\paragraph{Baseline Methods}
For AdamW, we conducted a grid search for the learning rate, and picked $2 \times 10^{-4}$ which performed the best.
We set $\beta_1$ to 0.9, $\beta_2$ to 0.999, and $\epsilon$ to $10^{-8}$.
For simplicity, we set the weight decay to 0 which is a common practice in LLM finetuning.
For Laplace-LoRA and BLoB, we use the same settings as in their code repositories, but we do not truncate the input context and do not finetune the output layer.

\paragraph{IVON-LoRA}
For IVON-LoRA, we set the learning rate to $5 \times 10^{-4}$, $\beta_1$ to 0.9, $\beta_2$ to 0.9998, and impose an element-wise clipping of preconditioned gradients to 0.1.
As shown by~\citet{shen2024variational}, the gradient clipping step is necessary for training language models.
We initialize $\vh$ to $5 \times 10^{-3}$.
For the effective training data size $\lambda$, we empirically set it to $5 \times 10^{5}$ for the small datasets (WG-S, ARC-E, and ARC-C) and $1 \times 10^{6}$ for the large datasets (WG-M, OBQA and BoolQ).

\subsection{Generation Tasks}
\label{app:gen}
\subsubsection{GSM8k}
We finetune Qwen2.5-3B~\citep{qwen2025qwen25technicalreport} on GSM8k~\citep{cobbe2021gsm8k} with the hyperparameters described in~\cref{subsec:hyperparameters}.
We do not use any quantization and train the model with an effective batch size of 16 on an NVIDIA A100 GPU with 40 GB memory.
We use a simple prompt prefix ``Solve the following math word problem: '' and do not use any in-context exemplars.
To perform prediction, we sample a varying number of outputs from the model using ancestral sampling and then perform uncertainty-aware MBR~\citep{daheim2025uncertaintyaware}.
To do so, we extract the final answer following the GSM8k output template and then use a 0-1-loss which reduces to majority voting over the final answer.
When using the IVON posterior we always sample 8 outputs from each model.
That is, for a total of 32 outputs we use 4 models to sample 8 times and for 64 outputs we sample 8 outputs from 8 models each and so on.
Empirically, this has performed better than using fewer models but more outputs in initial experiments.

\subsubsection{Conala}
We finetune Qwen2.5-3B~\citep{qwen2025qwen25technicalreport} on Conala~\citep{yin2018learning} with the hyperparameters described in~\cref{subsec:hyperparameters}.
Again, we do not use any quantization and train the model on a single NVIDIA A100 GPU with 40 GB memory.
We use uncertainty-aware MBR with 32 outputs in total, again using 8 samples from 4 models for the IVON posterior.
We use CodeBertScore~\citep{zhou-etal-2023-codebertscore} as a metric for MBR.
All considered model output samples are obtained with ancestral sampling.

\subsubsection{Hyperparameters}
\label{subsec:hyperparameters}
\paragraph{General Settings}
For all methods, we set the batch size to 16.
We set the number of training epochs to 6 and 10 for GSM8k and Conala, respectively, and use a linear learning rate scheduler.

\paragraph{Baseline Methods}
For the AdamW baseline, similar to the commonsense reasoning tasks, we set the learning rate to $2 \times 10^{-4}$, $\beta_1$ to 0.9, $\beta_2$ to 0.999, $\epsilon$ to $10^{-8}$ and weight decay to 0.

\paragraph{IVON-LoRA}
For IVON-LoRA, we set the learning rate to $3 \times 10^{-2}$, $\beta_1$ to 0.9, $\beta_2$ to 0.99999, and impose an element-wise clipping of preconditioned gradients to 0.01.
Furthermore, we initialize $\vh$ to $5 \times 10^{-3}$ and set $\lambda$ to $1 \times 10^{7}$.

\subsection{GLUE}
\label{app:glue}

We finetune DeBERTa-v3-base~\citep{he2021deberta} on the training set of the GLUE benchmark~\citep{wang2018glue} separately for each task.
Due to its large size and requirements for customized evaluation pipeline, we exclude the MNLI task from our evaluation, which will be included in the final version of the paper.
Both the base model and LoRA weights are kept in 16-bit precision.
Finetuning is performed on a single GeForce RTX 4090 GPU (with 24 GB memory).

\subsubsection{Hyperparameters}
\paragraph{General Settings}
Unlike in~\citet{hu2021lora}, we do not conduct extensive hyperparameter search, but instead use a fixed set of hyperparameters for all tasks.
We set the number of training epochs for each dataset to the minimum so that the number of epochs is no less than 3 and at least 5000 steps are taken.
For all methods, we set the batch size to 16.
We use a linear learning rate scheduler with 6\% of warmup steps.
Additionally, we truncate the input context to 512 tokens to reduce the memory footprint.

\paragraph{Baseline Methods}
For AdamW, we again set $\beta_1$ to 0.9, $\beta_2$ to 0.999, $\epsilon$ to $10^{-8}$ and weight decay to 0.
We conduct a grid search for the learning rate for both full-parameter and LoRA finetuning with AdamW.
For full-parameter and LoRA finetuning, we pick $1 \times 10^{-4}$ and $5 \times 10^{-4}$, respectively.

\paragraph{IVON-LoRA}
For IVON-LoRA, we set the learning rate to $1 \times 10^{-2}$, $\beta_1$ to 0.9, $\beta_2$ to 0.9998, and impose an element-wise clipping of preconditioned gradients to 0.02.
We initialize $\vh$ to $5 \times 10^{-3}$ and set $\lambda$ to $5 \times 10^{6}$.

\section{Licenses for Existing Assets}

\cref{tab:dataset} and \cref{tab:model} show the versions and licenses of the datasets and models used in this paper.

\begin{table}[h]
\centering
\begin{tabular}{lll}
\toprule
\textbf{Dataset} & \textbf{Version} & \textbf{License} \\
\midrule
WinoGrande              & 1.1.0 & CC-BY \\
ARC                     & 1.0.0 & CC-BY-SA-4.0 \\
OpenBookQA              & 1.0   & Apache-2.0 \\
BoolQ                   & 0.1.0 & CC-BY-SA-3.0 \\
GSM8K                   & 1.1.0 & MIT \\
CoNaLa                  & 1.1   & MIT \\
GLUE                    & 1.0.0 & CC-BY-SA-4.0 \\
\bottomrule
\end{tabular}
\caption{Versions and licenses of used datasets.}
\label{tab:dataset}
\end{table}

\begin{table}[h]
\centering
\begin{tabular}{ll}
\toprule
\textbf{Model} & \textbf{License} \\
\midrule
Llama-3.2-3B        & Llama 3.2 Community License \\
Qwen 2.5-3B         & Qwen Research License \\
DeBERTa-v3-base     & MIT \\
\bottomrule
\end{tabular}
\caption{Licenses of used models.}
\label{tab:model}
\end{table}

\end{document}